\pdfoutput=1

\documentclass[11pt]{article}

\usepackage{ACL2023}

\usepackage{times}
\usepackage{latexsym}
\usepackage{multirow}
\usepackage[T1]{fontenc}

\usepackage[utf8]{inputenc}

\usepackage{microtype}

\usepackage{inconsolata}
\usepackage{graphicx}
\graphicspath{ {images/} }

\title{TRAVID: An End-to-End Video Translation Framework}

\author{Prottay Kumar Adhikary \\
  NIT Silchar, Assam, India \\
  \texttt{prottay71@gmail.com} \\\And
  Sugandhi Bandaru \\
   NIT Silchar, Assam, India \\
  \texttt{bsugandhi99@gmail.com} \\\And
   Subhojit Ghimire\\
  NIT Silchar, Assam, India \\
  \texttt{subhojitg@live.com} \\
    \AND
  Santanu Pal \\
  Wipro AI Research (Lab45), London, UK \\
  \texttt{santanu.pal.ju@gmail.com} \\\And
  Partha Pakray \\
  NIT Silchar, Assam, India \\
  \texttt{parthapakray@gmail.com} \\
  }

\begin{document}
\maketitle
\begin{abstract}
In today's globalized world, effective communication with people from diverse linguistic backgrounds has become increasingly crucial. While traditional methods of language translation, such as written text or voice-only translations, can accomplish the task, they often fail to capture the complete context and nuanced information conveyed through nonverbal cues like facial expressions and lip movements. In this paper, we present an end-to-end video translation system that not only translates spoken language but also synchronizes the translated speech with the lip movements of the speaker. Our system focuses on translating educational lectures in various Indian languages, and it is designed to be effective even in low-resource system settings. By incorporating lip movements that align with the target language and matching them with the speaker's voice using voice cloning techniques, our application offers an enhanced experience for students and users. This additional feature creates a more immersive and realistic learning environment, ultimately making the learning process more effective and engaging.
\end{abstract}

\section{Introduction}
Face-to-Face (F2F) translation is a sub-field within the research domain of Machine Translation (MT). MT refers to the process of utilizing machines to translate text or speech from one language to another \cite{Somers92anintroduction}. F2F translation specifically focuses on translating spoken language in real-time during face-to-face conversations or interactions. The objective is to bridge language barriers and facilitate seamless communication between individuals who speak different languages.

F2F translation is also a part of the broader field of multi-modal machine translation, which integrates videos or visual information along with translation. This approach aims to enhance engagement among native language speakers during sessions. Visual cues, such as lip synchronization according to the native languages, contribute to a more realistic and immersive translated lecture session. These visual elements provide valuable context information that aids in the translation process.  Compared to image-guided multi-modal machine translation, videos provide visual and acoustic modalities with rich embedded information, such as actions, objects, and temporal transitions. From the past few years, image-based multi-modal models \citep{CHEN2022118264} only had marginal performance gains compared to their text-only counterparts, although very few of them are F2F translation \citep{CHEN2022108598}. 

F2F translation goes beyond traditional text-to-text or speech-to-speech translation methods. In a simple cascade-based F2F translation approach, several steps are involved: (i) \textbf{Capturing original speech:} the source video of a person delivering a speech is recorded or obtained, 
(ii) \textbf{Translating the captured speech:} the captured spoken language in the source video is translated to the desired language using machine translation techniques,
(iii) \textbf{Generating an output video:} Based on the translated text, an output video is generated where the same person appears to be speaking in the translated language, and 
(iv) \textbf{Maintaining lip synchronization:} during the generation of the output video, efforts are made to ensure that the lip movements of the person in the video match the target language, providing lip synchronization as per the translated language.
By following these steps, cascade-based F2F translation aims to deliver translated videos with synchronized lip movements, enhancing the authenticity and naturalness of the translated speech \citep{10.1145/3343031.3351066}. The intermediate steps i.e., \textbf{Translating the captured speech} can be modelled either direct \citep{app12031097} or cascade-based approach \citep{DBLP:journals/corr/abs-2011-12167}.  The cascade-based approach first performs a speech-to-text through an automatic speech recognizer (ASR) then the transcribed source text to desired target text using a text-to-text machine translation system and finally a text-to-speech system transform the translated text to speech in the desired language.

In addition to managing the individual components of our cascade-based system, we face significant challenges with F2F translation, particularly in the areas of lip synchronization and voice or tone alignment. The process involves recording a speech, converting it to text using speech-to-text technologies, translating this text from the original to the target language, and then converting the translated text back to speech via text-to-speech systems. This process can be achieved using either a cascade or direct approach. A major challenge in this end-to-end F2F translation framework is ensuring that the lip movements sync with the translated speech track. This can be complex, as the duration of the translated speech may be longer or shorter than the original, depending on the distinct grammatical structures of the two languages. Additionally, the lips must move in a manner consistent with the frequency of the generated sound and must maintain the speaker's original voice or tones. Failing to do so can result in dubbing that appears off and unrealistic \cite{10.1145/3394171.3413532}.

F2F translation can have a huge impact on bridging the language gap in the educational sector. Numerous educational organisations create content to reach a global audience. Unfortunately, the lack of language intelligibility often prevents content consumers from fully utilizing the material at hand. While some videos provide manually executed dubs, these have their own set of challenges. It's true that manual translation tends to be more accurate than machine translation, but it also faces unavoidable limitations. These include cost, availability, efficiency, and most importantly, the quality of lip synchronization, which often falls short of the mark \citep{10.1007/978-3-319-54427-4_19}. Additionally, manual dubbing may be available in many but not all languages.  The goal of the F2F translation system is to automate this dubbing process effectively and efficiently and make the online content available in whichever preferred language, thus overcoming the linguistic barrier between audio-visual content and the corresponding non-native consumer. This technology could also be used to assist language learning by giving students realistic and immersive opportunities to practise speaking and listening in a foreign language \citep{8682275}.
Through this paper, we contribute to creating a more equitable and accessible education landscape that enables native individuals to learn and grow without any language barrier. Our main objective is to motivate every individual by providing a platform, through which one can grasp knowledge from videos in an unfamiliar language. To the best of our knowledge, our F2F translation framework is the first online end-to-end video translation system we bring up to the community.


\section{Related Work}
\label{related}
In this section, we present part of previous studies conducted in this field and summarise our learning and inspiration to better complement our research.
\citet{prajwal2020learning} in their study explores the use of machine learning algorithms for lip-to-speech synthesis. The authors propose a new approach that takes into account individual speaking styles, resulting in increased accuracy. They use audio-visual data to train deep neural networks to capture unique lip movements and speaking styles, resulting in speech synthesis that is close to the original. The results show that their method outperforms current methods and produces speech that is similar to natural speech. \citet{10.1145/3343031.3351066} outlines a system for automatically translating speech between two people speaking different languages in real-time. The authors propose a multi-modal approach to translation that makes use of both audio and visual cues. This is accomplished by incorporating a novel visual module, LipGAN, for generating realistic talking faces in real-time from the translated audio. Their approach outperforms existing methods, demonstrating the potential for real-time F2F translation in practical applications. \citet{ritter1999face} in their research examines the development of a translation agent capable of performing real-time F2F speech translation. The authors present a multi-modal approach to translation that combines audio and visual information. They use machine learning algorithms to analyse each speaker's lip movements, speech, and facial expressions to produce a real-time audio-visual output with the speaker's face and synchronised lip movement. The results show that their method produces accurate translations and has the potential for practical applications in real-world scenarios. For translation, Chitralekha\footnote{\url{https://github.com/AI4Bharat/Chitralekha}} is a valuable tool because it efficiently creates multi-lingual subtitles and voice-overs for informative videos. However, it may not be a efficient for longer-length videos. Lastly, \citet{huang2017face} presented a novel problem of unpaired face translation between static photos and dynamic videos, which could be used to predict and improve video faces. To accomplish this task, the authors propose using a CycleGAN model with an identity-aware constraint. The model is trained on a large face dataset and tested on a variety of face images and videos. The results show that the proposed method can effectively translate faces between images and videos while preserving the individual's identity, outperforming existing methods.

\section{The TRAVID Framework}
\label{overview}
\begin{figure}[h]
\centering
\includegraphics[scale=0.4]{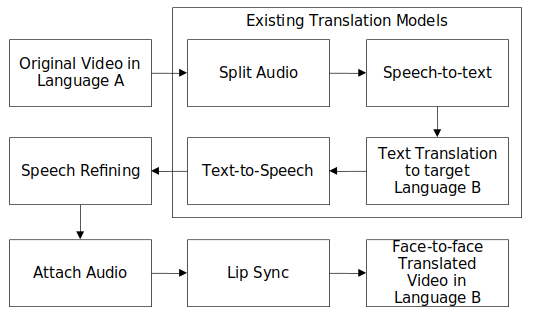}
\caption{\label{step}Steps involved in Video Translation System}
\end{figure}

Our framework `TRAVID' is capable of generating translated videos from English to four Indian languages: Bengali, Hindi, Nepali, and Telugu. Flask\footnote{\url{https://flask.palletsprojects.com/}} has been used as the foundation of our application, providing various built-in functionalities for building a Python-based web application. For the server-side and database, we utilize Python 3.9. In terms of audio and video processing, we primarily rely on the libraries Librosa\footnote{\url{https://librosa.org/doc/latest/index.html}} and ffmpeg\footnote{\url{https://pypi.org/project/ffmpeg-python/}}. These libraries provide extensive capabilities for audio and video processing, manipulation, and rendering. The primary objective of this work is to effectively and efficiently translate spoken language from an input video. Additionally, we aim to generate audio that resembles the speaker`s voice and synchronize the translated speech with the speaker's lip movements. The entire process begins by obtaining the source video, target language, and speaker's gender (for voice model selection) as input from the user through our web interface. Behind the scenes, the task is divided into three sub-tasks: (1) Audio-to-Text Processing, (2) Text-to-Audio Processing, and (3) Video Processing. The steps involved in this process are depicted in Figure \ref{step}.

\subsection{Audio to Text Processing}
The input video, in our case an MPEG-4 (.mp4) file, is initially converted to a Waveform Audio (.wav) file using FFmpeg. This conversion enables us to perform text detection from the audio rather than the video file. Subsequently, we employ Librosa to identify non-mute sections within the `start' and `end' frame indexes, which are stored in a silence array. Each element of the silence array represents a small audio chunk, aiding in reducing system load and enhancing the overall efficiency of the framework during audio processing. Next, we convert each audio chunk from the silence array into an individual text chunk using Speech Recognition\footnote{\url{https://pypi.org/project/SpeechRecognition/}}. This library utilizes Google's Cloud speech API\footnote{\url{www.pypi.org/project/google-cloud-speech}} to covert text from speech. Finally, Deep-translator\footnote{\url{https://pypi.org/project/deep-translator/}} is employed to translate the generated text into the target language. Deep Translator utilizes the state-of-the-art Google Translate Ajax API\footnote{\url{https://pypi.org/project/googletrans/}} to generate the desired target language translation. The translated texts are stored and subsequently passed to the audio speech engine for further processing.


\subsection{Text to Audio Processing}
The translated text in the target language is inserted to the gTTS\footnote{\url{https://pypi.org/project/gTTS/}} library, which converts the text into speech and saves it as an audio file. This marks the completion of the speech generation process and initiates the speech refinement process. In order to match the audio target length with the source audio length, adjustments are made. The length of the translated speech may differ from that of the original speech. To address this, the speech speed is modified to align with the original audio file. The ``Fixed Pitch-Shifting'' technique is employed to ensure that the generated speech closely resembles the voice of the original speaker. Librosa provides the capability to detect the frequency of the audio and shift the pitch of the audio time series from one musical note to another \citep{rosenzweig2021adaptive}. In the context of voice cloning, the mean frequency of the audio is determined, with the lower note considered as F2 (87.31 Hz) and the higher note as G6 (98.00 Hz). This frequency range represents the average range of human speech. The calculation of the steps required for shifting ($n\_steps$) is performed using Equation \ref{eq:nstep}.
\begin{equation}
\small
\label{eq:nstep}
    n\_steps=\log_{2}(\frac{f_{src}}{f_{tgt}})^{2}
\end{equation}

The variable $f_{src}$ refers to the frequency of the source audio and $f_{tgt}$ refers to the corresponding target audio. With this, the Text to Audio Processing engine gives the desired audio to the video-processing engine.

\subsection{Video Processing for Lip Synchronization}
We have utilized a lip-synchronization network called Wav2Lip \citep{10.1145/3394171.3413532} for the purpose of lip-syncing and generating talking face videos. This model has been trained on the LRS2 training set and demonstrates an approximate accuracy of 91\% on the LRS2 test set. The video sub-network of the model examines each frame of the source video and identifies faces, with a particular focus on the lip region. The relevant audio segment is then fed into the speech sub-network component of Wav2Lip, which modifies the input face crop to emphasize the lips area and produces the final video output. Throughout this process, the lip portion of the source video is replaced by concatenating the current face crop with the lower half of the detected face. By leveraging the translated speech and the source video, Wav2Lip generates lip-synced translated videos. The resulting translated video is subsequently presented on our front-end for display.

\section{Demo Scenarios}
\label{demosec}
\begin{figure}[h]
\centering
\includegraphics[scale=0.66]{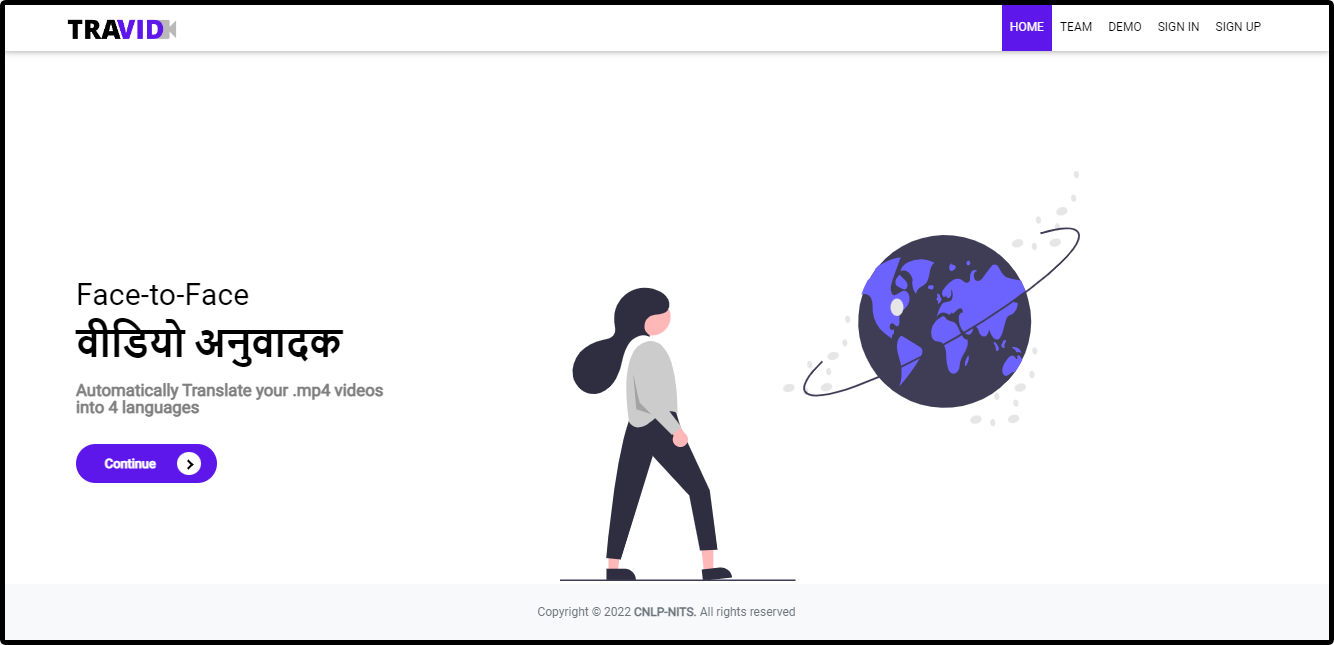}
\caption{\label{home}Homepage of TRAVID}
\end{figure}

Our framework TRAVID has a visually appealing landing page\footnote{\url{https://github.com/human71/TRAVID}}, which has an overview of the framework (cf. Figure \ref{home}). 
A demonstration video of our system is available on YouTube\footnote{\url{https://youtu.be/XNNp1xF5H0Y}}. 
The demo User Interface (UI) is designed a landing page, have been carefully crafted to provide a seamless and intuitive navigation experience.
The landing page is effectively communicates its purpose and functionality without the need for extensive instructions or guidance. Users can easily understand what the page offers and how to navigate it intuitively. The page is organized into distinct sections that make it easy for users to locate and access the information they are looking for. This organization achieved through the use of clear headings, visually distinct sections, or a logical flow of content. The top menu bar is mentioned as a key element of the landing page, providing menu options that direct users to different feature pages. This menu bar remains accessible and visible to users across different sections of the landing page, allowing them to navigate to specific areas of interest easily.
Upon signing in or signing up as a new user, the statement states that the user will be directed to the core section of the demo. This core section is the central part of the landing page, where users can access the main functionality and key features of TRAVID. 

\begin{figure}[h]
\centering
\includegraphics[scale=0.66]{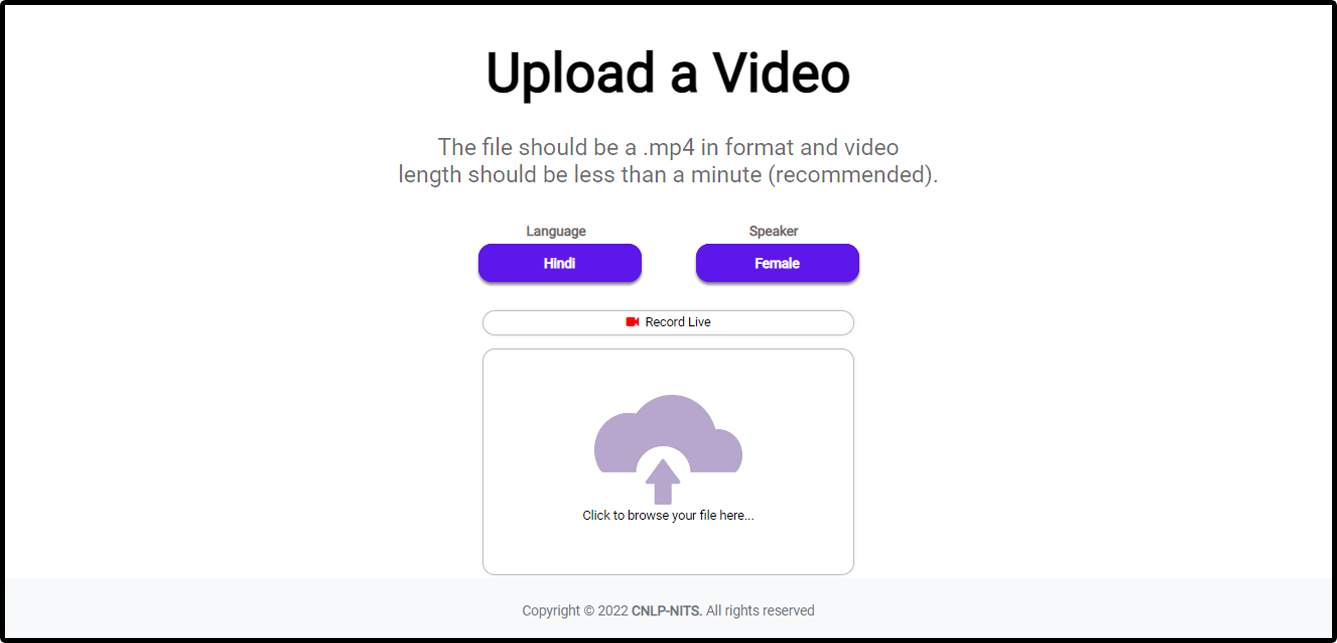}
\caption{\label{upload}Upload page of TRAVID}
\end{figure}

The upload page shown in Figure \ref{upload} includes two drop-down menus: one for selecting the desired language for translation and another for choosing the output voice model (speaker). There are two options available for video input on the upload page: live recording, which allows users to capture real-time audio-visual input using their device's camera and microphone, or accepting pre-saved audio-visual content from the system. After receiving the input, the back-end framework, discussed in Section \ref{overview}, initiates the translation process. Once the text, audio, and video processing are complete, the output page displays the translated video alongside the source video. Users have the option to download the source video and translated text. Additionally, they can provide reviews based on the output they received, which can help us improve and enhance the user-friendliness of our system.

\begin{figure}[h]
\centering
\includegraphics[scale=0.66]{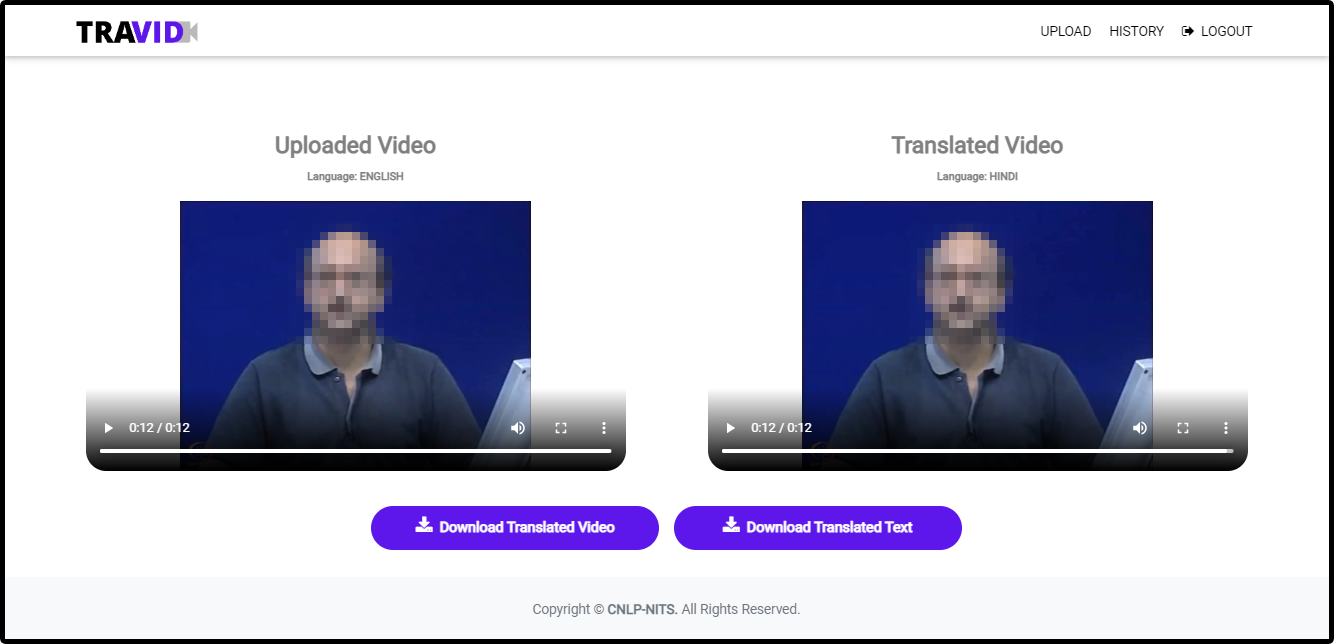}
\caption{\label{output}Output page of TRAVID}
\end{figure}

\begin{figure}[h]
\centering
\includegraphics[scale=0.66]{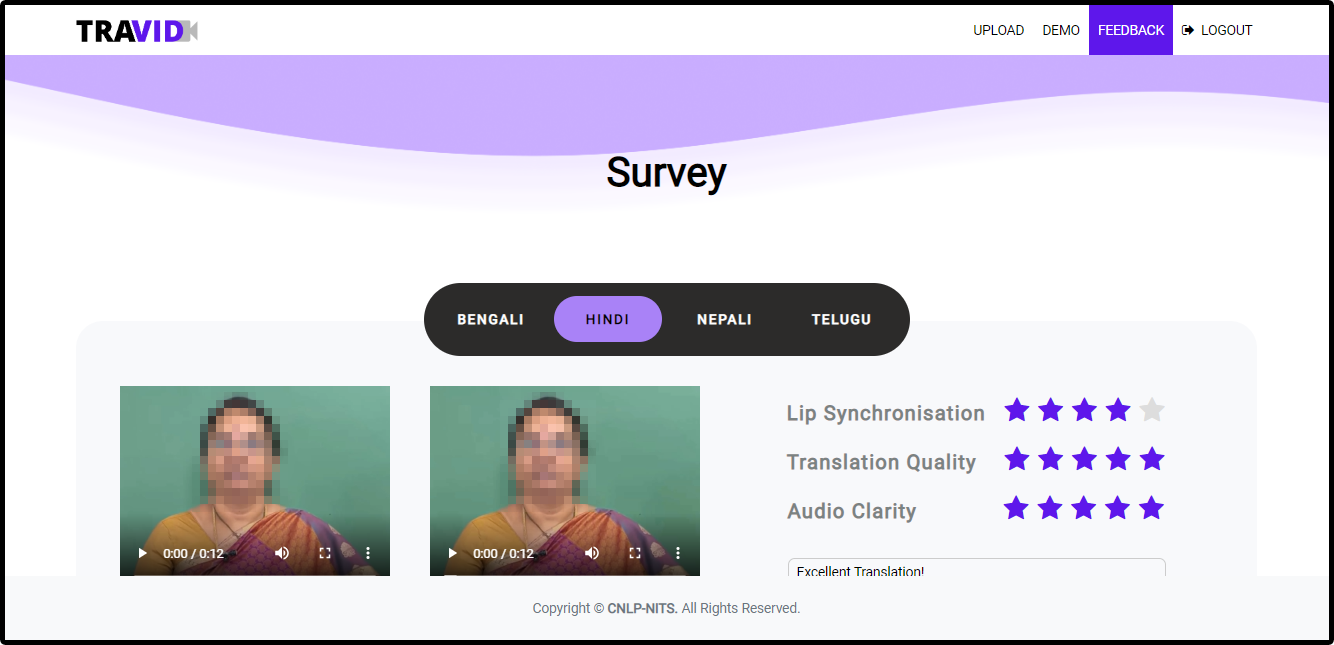}
\caption{\label{survey}Survey page of TRAVID}
\end{figure}

The output page, depicted in Figure \ref{output}, provides a clear presentation of the original input video and the generated output video side-by-side. It offers convenient options to play and review both videos simultaneously. Additionally, users can save and download both the translated video and a translated text document.
Furthermore, users can explore the demo section, which displays test case videos, to gain an understanding of the translation quality that TRAVID's translation model can produce. In addition to the demo section, TRAVID includes a feedback page where users can rate videos alongside their translated output according to specific criteria and provide feedback to enhance our framework (refer to Figure \ref{survey}).

\section{Evaluation}
\label{comp}
To gauge the effectiveness of our method, 
We conducted a user study to assess the quality of our lip-synced translations, with participants asked to rate the translation quality, lip synchronization, and audio clarity. Evaluators compared the target video with the source language video clip and provided rankings for the quality of the output video on a scale of 1 to 5. The collected ratings were used to calculate inter-annotator agreement using Cohen's $\kappa$ \cite{Cohen1960ACO}, Fleiss' $\kappa$ \cite{fleiss1971measuring}, and Pearson's $r$ \cite{1895RSPS...58..240P} scores. Inter-annotator agreements were computed for all four languages: Bengali, Hindi, Nepali, and Telugu. Table \ref{tab:agreement} displays the agreement scores for each language based on Lip Synchronization (Lip Sync), Translation Quality (TQ), and Audio Quality (AQ).
The ratings were collected by comparing the translated videos to source videos from 5 indigenous users for each of the selected languages. Moreover, a manual examination was conducted by professional evaluators, and the results are presented in Table \ref{tab:iitmeval}. Further details regarding inter-rater agreements can be found in Appendix \ref{sec:app:cohen}, \ref{sec:app:fkappa}, \ref{sec:app:pearson}.

The core component of TRAVID is based on the CNLP\_NITS system, which emerged as a top performer in the Lip-Sync 2021 Challenge shared task\footnote{\href{https://nplt.in/demo/leadership-board?fbclid=IwAR1uNyvjB6zvXKOqyFtFXVdPcgzPqEzQ25xFsLItYvUIQW0v4EzSBU-UZuw}{Leaderboard of Lip-Sync Challenge 2021}}. The objective of this challenge was to convert English input videos into Hindi or Tamil output videos while ensuring lip synchronization. The quality of the Hindi Task-1 was assessed using various evaluation metrics such as Lip-Sync Quality (LSQ), Fluency Consistency (FC), Semantic Consistency (SC), and Overall User Experience (UX). Evaluators rated the quality of the output videos on a scale of 1 to 5, with higher scores indicating better quality when compared to the source language video clip. Our system, CNLP-NITS (NIT Silchar), achieved the top position with a final score of 3.84, surpassing the Baseline system (IIT Madras) with a score of 3.68 and TeamCSRL (CS RESEARCH LABS) with a score of 3.46. The comparison of the three evaluation matrices revealed a high degree of similarity. The results indicate that the translations were perceived as reasonable and easy to understand by the majority of participants, leading to fair to moderate agreement and a positive correlation among their assessments. The overall scores of the Lip-Sync Challenge 2021 are presented in Table \ref{tab:iitmeval}.

\begin{table}[!htb]
\centering
\small
\begin{tabular}{ccc|c}

\hline
\multicolumn{3}{c|}{Hindi Task 1}                                                             & Hindi Task 2                  \\ \hline
\multicolumn{1}{l}{CNLP\_NITS} & \multicolumn{1}{l}{Baseline} & \multicolumn{1}{l|}{TeamCSRL} & \multicolumn{1}{l}{CNLP-NITS} \\ \hline
3.86                           & 3.49                         & 3.08                          & 3.37                          \\
3.63                           & 3.52                         & 3.32                          & 3.29                          \\
3.94                           & 3.87                         & 3.92                          & 3.40                           \\
3.94                           & 3.83                         & 3.51                          & 3.38                          \\
3.84                           & 3.68                         & 3.46                          & 3.36                          \\ \hline
\end{tabular}

\caption{\label{tab:iitmeval}Leadership Positions Based on NLP Challenges}
\end{table}

\begin{table*}[t]
 \small
\centering
\begin{tabular}{l|ccc|ccc|ccc}
\hline
\multicolumn{1}{c|}{\multirow{2}{*}{\textbf{Language}}} & \multicolumn{3}{c|}{\textbf{Cohen's $\kappa$}}          & \multicolumn{3}{c|}{\textbf{Fleiss' $\kappa$}} & \multicolumn{3}{c}{\textbf{Pearson's $r$}} \\
\cline{2-10}
\multicolumn{1}{c|}{}                                   & \textbf{Lip Sync} & \textbf{TQ} & \textbf{AQ} & \textbf{Lip Sync}  & \textbf{TQ} & \textbf{AQ} & \textbf{Lip Sync}  & \textbf{TQ}  & \textbf{AQ} \\
\hline
\textbf{Bengali}                                       & 0.600          & 0.276    & 0.296    & 0.698  & 0.379    & 0.215    & 0.586  & 0.295     & 0.258    \\
\textbf{Hindi}                                         & 0.510          & 0.390    & 0.391    & 0.595  & 0.291    & 0.345    & 0.400  & 0.292     & 0.318    \\
\textbf{Nepali}                                        & 0.170          & 0.576    & 0.328    & 0.112  & 0.541    & 0.218    & 0.171  & 0.501     & 0.256    \\
\textbf{Telugu}                                        & 0.287          & 0.330    & 0.388    & 0.214  & 0.291    & 0.376    & 0.212  & 0.271     & 0.331  \\
\hline
\end{tabular}
\caption{\label{tab:agreement} Average agreement Scores for evaluation of TRAVID generated Videos}
\end{table*}

\section{Limitation}
There is a constraint when uploading huge videos; the system may require a lot of computational resources and data to process and render the translated video. Also, so far we have trained our models only on a single speaker, so videos with multiple speakers may yield poor results. The quality of speech recognition and translation may vary depending on factors such as noise, accent, dialect, etc. The generated faces may not look natural or convincing enough for some applications or scenarios such as low lighting, moving background, etc. Considering the state-of-the-art ASR system in use, the ASR results were already deemed satisfactory, thus not necessitating the utilization of lip sync from the video as an additional multimodal input for accuracy enhancement. Still, the system may be unable to handle linguistic challenges such as idioms, metaphors, slang, etc. The method may not be able to capture cultural nuances and context that affect the meaning and tone of speech, as the syntheses is machine generated. The biggest bottleneck in our current system which  uses cascade approach is time complexity, due to the need for extensive computation and audiovisual processing.

\section{Conclusion}
\label{end}

In this paper, we presented an end-to-end video translation system that effectively translates the speaker's native language into the local language of the audience while synchronizing the translated speech with the speaker's lip movements. Our proposed system demonstrates the potential of lip-synced Face-to-Face video translation in enhancing communication between individuals from diverse linguistic backgrounds.

Moreover, our video translation system represents a significant advancement in overcoming the limitations of traditional language translation methods. By incorporating lip synchronization and matching translated speech with the speaker's lip movements, we created an immersive and realistic experience for users. This additional feature, along with the ability to capture nonverbal cues, adds depth and context to the translated content, making it more effective and engaging, especially in educational settings.

Through our system's participation and success in the Lip-Sync 2021 Challenge, we have demonstrated its capability and superiority in achieving accurate lip synchronization and high-quality translations. The evaluations and ratings obtained from both users and professional evaluators validate the effectiveness of our approach, further emphasizing its potential for real-world applications. The positive feedback received through human assessments, as discussed in the evaluation section above, validates the effectiveness of our system. However, further research is necessary to enhance the quality of lip-syncing and explore the system's applicability in different languages and more naturalistic settings. With the ongoing advancements in technology and the increasing demand for multilingual communication, our system has the potential to revolutionize the way language translation is approached. Its adaptability to low-resource system settings makes it accessible and valuable in diverse environments.


Moving forward, we envision further enhancements and refinements to our video translation system, leveraging the advancements in natural language processing, computer vision, and machine learning. To boost video translation efficiency, the videos can be broken into smaller segments, leverage GPUs for parallel processing, batch translate frames, subsample for reduced load, and implement caching for reused translations. By continuously improving the accuracy, fluency, and naturalness of translated content, we aim to provide an unparalleled experience for users, fostering effective cross-cultural communication and knowledge sharing.

In summary, our video translation system stands as a promising solution to the challenges of multilingual communication, offering a comprehensive and immersive experience that unlocks new possibilities for global connectivity and understanding.


\section*{Ethics Statement}
We honour the Code of Ethics set by IJCNLP-AACL in our paper and abide by them. We have used open-source materials in our development to produce new, better and useful resources, which will be made open-source for the keen mind to feed upon and make more improvements in the future. We have not written or in any way propagated false knowledge, hateful speech and anything controversial that may give rise to conflict. We intend good for the brighter future of mankind. We have not stolen anybody's work, and have properly cited and credited where credit is due. Our website does not feature any harmful content or advertisement. Our website is solely educational and is to be used for educational purposes only. Even though we reserve the right to use our paper and production in any way we see fit, we promise to extend them ethically and in an innovative manner.

\bibliography{acl2023}

\begin{thebibliography}{16}
\expandafter\ifx\csname natexlab\endcsname\relax\def\natexlab#1{#1}\fi

\bibitem[{Bahar et~al.(2020)Bahar, Bieschke, Schl{\"{u}}ter, and Ney}]{DBLP:journals/corr/abs-2011-12167}
Parnia Bahar, Tobias Bieschke, Ralf Schl{\"{u}}ter, and Hermann Ney. 2020.
\newblock \href {http://arxiv.org/abs/2011.12167} {Tight integrated end-to-end training for cascaded speech translation}.
\newblock \emph{CoRR}, abs/2011.12167.

\bibitem[{Chen et~al.(2022{\natexlab{a}})Chen, Zeng, Cao, and Lu}]{CHEN2022108598}
Shiyu Chen, Yawen Zeng, Da~Cao, and Shaofei Lu. 2022{\natexlab{a}}.
\newblock \href {https://doi.org/https://doi.org/10.1016/j.knosys.2022.108598} {Video-guided machine translation via dual-level back-translation}.
\newblock \emph{Knowledge-Based Systems}, 245:108598.

\bibitem[{Chen et~al.(2022{\natexlab{b}})Chen, Zeng, Cao, and Lu}]{CHEN2022118264}
Shiyu Chen, Yawen Zeng, Da~Cao, and Shaofei Lu. 2022{\natexlab{b}}.
\newblock \href {https://doi.org/https://doi.org/10.1016/j.eswa.2022.118264} {Vision talks: Visual relationship-enhanced transformer for video-guided machine translation}.
\newblock \emph{Expert Systems with Applications}, 209:118264.

\bibitem[{Chung and Zisserman(2017)}]{10.1007/978-3-319-54427-4_19}
Joon~Son Chung and Andrew Zisserman. 2017.
\newblock Out of time: Automated lip sync in the wild.
\newblock In \emph{Computer Vision -- ACCV 2016 Workshops}, pages 251--263, Cham. Springer International Publishing.

\bibitem[{Cohen(1960)}]{Cohen1960ACO}
Jacob Cohen. 1960.
\newblock A coefficient of agreement for nominal scales.
\newblock \emph{Educational and Psychological Measurement}, 20:37 -- 46.

\bibitem[{Etchegoyhen et~al.(2022)Etchegoyhen, Arzelus, Gete, Alvarez, Torre, Martín-Doñas, González-Docasal, and Fernandez}]{app12031097}
Thierry Etchegoyhen, Haritz Arzelus, Harritxu Gete, Aitor Alvarez, Iván~G. Torre, Juan~Manuel Martín-Doñas, Ander González-Docasal, and Edson~Benites Fernandez. 2022.
\newblock \href {https://doi.org/10.3390/app12031097} {Cascade or direct speech translation? a case study}.
\newblock \emph{Applied Sciences}, 12(3).

\bibitem[{Fleiss(1971)}]{fleiss1971measuring}
Joseph~L Fleiss. 1971.
\newblock Measuring nominal scale agreement among many raters.
\newblock \emph{Psychological bulletin}, 76(5):378.

\bibitem[{Huang et~al.(2017)Huang, Kratzwald, Paudel, Wu, and Van~Gool}]{huang2017face}
Zhiwu Huang, Bernhard Kratzwald, Danda~Pani Paudel, Jiqing Wu, and Luc Van~Gool. 2017.
\newblock Face translation between images and videos using identity-aware cyclegan.
\newblock \emph{arXiv preprint arXiv:1712.00971}.

\bibitem[{Jha et~al.(2019)Jha, Voleti, Namboodiri, and Jawahar}]{8682275}
Abhishek Jha, Vikram Voleti, Vinay Namboodiri, and C.~V. Jawahar. 2019.
\newblock \href {https://doi.org/10.1109/ICASSP.2019.8682275} {Cross-language speech dependent lip-synchronization}.
\newblock In \emph{ICASSP 2019 - 2019 IEEE International Conference on Acoustics, Speech and Signal Processing (ICASSP)}, pages 7140--7144.

\bibitem[{K~R et~al.(2019)K~R, Mukhopadhyay, Philip, Jha, Namboodiri, and Jawahar}]{10.1145/3343031.3351066}
Prajwal K~R, Rudrabha Mukhopadhyay, Jerin Philip, Abhishek Jha, Vinay Namboodiri, and C~V Jawahar. 2019.
\newblock \href {https://doi.org/10.1145/3343031.3351066} {Towards automatic face-to-face translation}.
\newblock In \emph{Proceedings of the 27th ACM International Conference on Multimedia}, MM '19, page 1428–1436, New York, NY, USA. Association for Computing Machinery.

\bibitem[{{Pearson}(1895)}]{1895RSPS...58..240P}
Karl {Pearson}. 1895.
\newblock {Note on Regression and Inheritance in the Case of Two Parents}.
\newblock \emph{Proceedings of the Royal Society of London Series I}, 58:240--242.

\bibitem[{Prajwal et~al.(2020{\natexlab{a}})Prajwal, Mukhopadhyay, Namboodiri, and Jawahar}]{10.1145/3394171.3413532}
K~R Prajwal, Rudrabha Mukhopadhyay, Vinay~P. Namboodiri, and C.V. Jawahar. 2020{\natexlab{a}}.
\newblock \href {https://doi.org/10.1145/3394171.3413532} {A lip sync expert is all you need for speech to lip generation in the wild}.
\newblock In \emph{Proceedings of the 28th ACM International Conference on Multimedia}, MM '20, page 484–492, New York, NY, USA. Association for Computing Machinery.

\bibitem[{Prajwal et~al.(2020{\natexlab{b}})Prajwal, Mukhopadhyay, Namboodiri, and Jawahar}]{prajwal2020learning}
KR~Prajwal, Rudrabha Mukhopadhyay, Vinay~P Namboodiri, and CV~Jawahar. 2020{\natexlab{b}}.
\newblock Learning individual speaking styles for accurate lip to speech synthesis.
\newblock In \emph{Proceedings of the IEEE/CVF Conference on Computer Vision and Pattern Recognition}, pages 13796--13805.

\bibitem[{Ritter et~al.(1999)Ritter, Meier, Yang, and Waibel}]{ritter1999face}
Max Ritter, Uwe Meier, Jie Yang, and Alex Waibel. 1999.
\newblock Face translation: A multimodal translation agent.
\newblock In \emph{AVSP'99-International Conference on Auditory-Visual Speech Processing}.

\bibitem[{Rosenzweig et~al.(2021)Rosenzweig, Schw{\"a}r, Driedger, and M{\"u}ller}]{rosenzweig2021adaptive}
Sebastian Rosenzweig, Simon Schw{\"a}r, Jonathan Driedger, and Meinard M{\"u}ller. 2021.
\newblock Adaptive pitch-shifting with applications to intonation adjustment in a cappella recordings.
\newblock In \emph{2021 24th International Conference on Digital Audio Effects (DAFx)}, pages 121--128. IEEE.

\bibitem[{Somers(1992)}]{Somers92anintroduction}
Harold Somers. 1992.
\newblock An introduction to machine translation.

\end{thebibliography}
\bibliographystyle{acl_natbib}
\clearpage
\appendix

\section{Cohen's Kappa}
\label{sec:app:cohen}
The definition of  $\kappa$ is
\[
\kappa = \frac{{p_o}- {p_e}}{1-{p_e}}
\]
where ${p_o}$ denotes the relative observed agreement among raters and $p_e$ denotes the hypothetical probability of chance agreement, calculated by using observed data to compute the odds of each observer randomly seeing each group.

\section{Fleiss’ Kappa}
\label{sec:app:fkappa}
The Fleiss’ kappa $\kappa$ can be defined as,
\[
\kappa = \frac{\bar{P}- \bar{P_e}}{1-\bar{P_e}}
\]
The factors  $1-{\bar  {P_{e}}}$  and ${\bar  {P}}-{\bar  {P_{e}}}$ indicate the level of agreement that can be reached above chance and the level of agreement that has actually been attained.

\section{Pearson Correlation Coefficient Analysis}
\label{sec:app:pearson}
Given a pair of random variables $(X,Y)$, the formula for $\rho$ is:
\[
\rho_{X,Y} = \frac{cov(X,Y)}{\sigma_X \sigma_Y}
\]
where $cov$ is the covariance, $\sigma_X$ is the standard deviation of $X$ and $\sigma_Y$ is the standard deviation of $Y$.\\
Table [\ref{1}, \ref{2}, \ref{3}] displays the calculated Pearson Correlation Coefficient (PCC) values for Bengali, Table [\ref{4}, \ref{5}, \ref{6}] displays the calculated PCC values for Hindi, Table [\ref{7}, \ref{8}, \ref{9}] displays the calculated PCC values for Nepali, Table [\ref{10}, \ref{11}, \ref{12}] displays the calculated PCC values for Telugu with respect to all three assessment areas, namely Lip Synchronization, Translation Quality, and Audio Quality. Where R1-R5 (Rater 01-05) denote five raters who assessed our system in all four languages.\\
\begin{table}[!htb]
\centering
\small
\begin{tabular}{lrrrrr}
                        & \multicolumn{1}{l}{R1} & \multicolumn{1}{l}{R2} & \multicolumn{1}{l}{R3} & \multicolumn{1}{l}{R4} & \multicolumn{1}{l}{R5} \\ \cline{2-6} 
\multicolumn{1}{l|}{R1} & 1                      & -0.2                   & 0.81                   & 0.72                   & 0.46                   \\
\multicolumn{1}{l|}{R2} & -0.2                   & 1                      & 0.75                   & 0.81                   & 0.85                   \\
\multicolumn{1}{l|}{R3} & 0.81                   & 0.75                   & 1                      & 0.84                   & -0.02                  \\
\multicolumn{1}{l|}{R4} & 0.72                   & 0.55                   & 0.84                   & 1                      & 0.84                   \\
\multicolumn{1}{l|}{R5} & 0.46                   & 0.85                   & -0.02                  & 0.84                   & 1                     
\end{tabular}
\caption{\label{1}PCC for Bengali Lip Synchronization}
\end{table}

\begin{table}[!htb]
\centering
\small
\begin{tabular}{lrrrrr}
                        & \multicolumn{1}{l}{R1} & \multicolumn{1}{l}{R2} & \multicolumn{1}{l}{R3} & \multicolumn{1}{l}{R4} & \multicolumn{1}{l}{R5} \\ \cline{2-6} 
\multicolumn{1}{l|}{R1} & 1                      & 0.32                   & 0.86                   & 0.21                   & 0.87                   \\
\multicolumn{1}{l|}{R2} & 0.32                   & 1                      & 0.68                   & 0.12                   & 0.38                   \\
\multicolumn{1}{l|}{R3} & 0.86                   & 0.68                   & 1                      & -0.64                  & 0.57                   \\
\multicolumn{1}{l|}{R4} & 0.21                   & 0.12                   & -0.64                  & 1                      & -0.42                  \\
\multicolumn{1}{l|}{R5} & 0.87                   & 0.38                   & 0.57                   & -0.42                  & 1                     
\end{tabular}
\caption{\label{2}PCC for Bengali Translation Quality}
\end{table}

\begin{table}[!htb]
\centering
\small
\begin{tabular}{lrrrrr}
                        & \multicolumn{1}{l}{R1} & \multicolumn{1}{l}{R2} & \multicolumn{1}{l}{R3} & \multicolumn{1}{l}{R4} & \multicolumn{1}{l}{R5} \\ \cline{2-6} 
\multicolumn{1}{l|}{R1} & 1                      & 0.79                   & -0.18                  & 0.25                   & 0.33                   \\
\multicolumn{1}{l|}{R2} & 0.79                   & 1                      & 0.42                   & 0.39                   & 0.12                   \\
\multicolumn{1}{l|}{R3} & -0.18                  & 0.42                   & 1                      & -0.19                  & 0.39                   \\
\multicolumn{1}{l|}{R4} & 0.25                   & 0.39                   & -0.19                  & 1                      & 0.26                   \\
\multicolumn{1}{l|}{R5} & 0.33                   & 0.12                   & 0.39                   & 0.26                   & 1                     
\end{tabular}
\caption{\label{3}PCC for Bengali Audio Quality}
\end{table}

\begin{table}[!htb]
\centering
\small
\begin{tabular}{lrrrrr}
                        & \multicolumn{1}{l}{R1} & \multicolumn{1}{l}{R2} & \multicolumn{1}{l}{R3} & \multicolumn{1}{l}{R4} & \multicolumn{1}{l}{R5} \\ \cline{2-6} 
\multicolumn{1}{l|}{R1} & 1                      & -0.12                  & 0.26                   & -0.06                  & 0.21                   \\
\multicolumn{1}{l|}{R2} & -0.12                  & 1                      & 0.55                   & 0.39                   & 0.65                   \\
\multicolumn{1}{l|}{R3} & 0.26                   & 0.55                   & 1                      & 0.66                   & 0.72                   \\
\multicolumn{1}{l|}{R4} & -0.06                  & 0.39                   & 0.66                   & 1                      & 0.74                   \\
\multicolumn{1}{l|}{R5} & 0.21                   & 0.65                   & 0.72                   & 0.74                   & 1                     
\end{tabular}
\caption{\label{4}PCC for Hindi Lip Synchronization}
\end{table}

\begin{table}[!htb]
\centering
\small
\begin{tabular}{lrrrrr}
                        & \multicolumn{1}{l}{R1} & \multicolumn{1}{l}{R2} & \multicolumn{1}{l}{R3} & \multicolumn{1}{l}{R4} & \multicolumn{1}{l}{R5} \\ \cline{2-6} 
\multicolumn{1}{l|}{R1} & 1                      & -0.42                  & -0.42                  & 0.51                   & 0.66                   \\
\multicolumn{1}{l|}{R2} & -0.42                  & 1                      & 0.88                   & 0.56                   & -0.12                  \\
\multicolumn{1}{l|}{R3} & 0.36                   & 0.88                   & 1                      & 0.76                   & 0.62                   \\
\multicolumn{1}{l|}{R4} & 0.51                   & 0.56                   & 0.76                   & 1                      & -0.11                  \\
\multicolumn{1}{l|}{R5} & 0.66                   & -0.12                  & 0.62                   & -0.11                  & 1                     
\end{tabular}
\caption{\label{5}PCC for Hindi Translation Quality}
\end{table}

\begin{table}[!htb]
\centering
\small
\begin{tabular}{lrrrrr}
                        & \multicolumn{1}{l}{R1} & \multicolumn{1}{l}{R2} & \multicolumn{1}{l}{R3} & \multicolumn{1}{l}{R4} & \multicolumn{1}{l}{R5} \\ \cline{2-6} 
\multicolumn{1}{l|}{R1} & 1                      & 0.22                   & 0.68                   & -0.01                  & 0.72                   \\
\multicolumn{1}{l|}{R2} & 0.22                   & 1                      & -0.19                  & 0.3                    & 0.59                   \\
\multicolumn{1}{l|}{R3} & 0.68                   & -0.19                  & 1                      & 0.12                   & 0.26                   \\
\multicolumn{1}{l|}{R4} & -0.01                  & 0.3                    & 0.12                   & 1                      & 0.49                   \\
\multicolumn{1}{l|}{R5} & 0.72                   & 0.59                   & 0.26                   & 0.49                   & 1                     
\end{tabular}
\caption{\label{6}PCC for Hindi Audio Quality}
\end{table}

\begin{table}[!htb]
\centering
\small
\begin{tabular}{lrrrrr}
                        & \multicolumn{1}{l}{R1} & \multicolumn{1}{l}{R2} & \multicolumn{1}{l}{R3} & \multicolumn{1}{l}{R4} & \multicolumn{1}{l}{R5} \\ \cline{2-6} 
\multicolumn{1}{l|}{R1} & 1                      & -0.2                   & -0.01                  & 0.52                   & 0.11                   \\
\multicolumn{1}{l|}{R2} & -0.2                   & 1                      & 0.16                   & 0.29                   & 0.21                   \\
\multicolumn{1}{l|}{R3} & -0.01                  & 0.16                   & 1                      & -0.22                  & 0.34                   \\
\multicolumn{1}{l|}{R4} & 0.52                   & 0.29                   & -0.22                  & 1                      & 0.51                   \\
\multicolumn{1}{l|}{R5} & 0.11                   & 0.21                   & 0.34                   & 0.51                   & 1                     
\end{tabular}
\caption{\label{7}PCC for Nepali Lip Synchronization}
\end{table}

\begin{table}[!htb]
\centering
\small
\begin{tabular}{lrrrrr}
                        & \multicolumn{1}{l}{R1} & \multicolumn{1}{l}{R2} & \multicolumn{1}{l}{R3} & \multicolumn{1}{l}{R4} & \multicolumn{1}{l}{R5} \\ \cline{2-6} 
\multicolumn{1}{l|}{R1} & 1                      & 0.69                   & 0.58                   & -0.09                  & 0.66                   \\
\multicolumn{1}{l|}{R2} & 0.69                   & 1                      & 0.49                   & 0.63                   & -0.21                  \\
\multicolumn{1}{l|}{R3} & 0.58                   & 0.49                   & 1                      & 0.66                   & 0.79                   \\
\multicolumn{1}{l|}{R4} & -0.09                  & 0.63                   & 0.66                   & 1                      & 0.81                   \\
\multicolumn{1}{l|}{R5} & 0.66                   & -0.21                  & 0.79                   & 0.81                   & 1                     
\end{tabular}
\caption{\label{8}PCC for Nepali Translation Quality}
\end{table}
\clearpage
\begin{table}[!htb]
    \begin{minipage}{0.9\linewidth}
      \centering
      \small
        \begin{tabular}{lrrrrr}
                        & \multicolumn{1}{l}{R1} & \multicolumn{1}{l}{R2} & \multicolumn{1}{l}{R3} & \multicolumn{1}{l}{R4} & \multicolumn{1}{l}{R5} \\ \cline{2-6} 
\multicolumn{1}{l|}{R1} & 1                      & 0.11                   & -0.02                  & -0.15                  & 0.29                   \\
\multicolumn{1}{l|}{R2} & 0.11                   & 1                      & 0.59                   & 0.63                   & 0.82                   \\
\multicolumn{1}{l|}{R3} & -0.02                  & 0.59                   & 1                      & 0.26                   & -0.09                  \\
\multicolumn{1}{l|}{R4} & -0.15                  & 0.63                   & 0.26                   & 1                      & 0.12                   \\
\multicolumn{1}{l|}{R5} & 0.29                   & 0.82                   & -0.09                  & 0.12                   & 1                     
\end{tabular}
\caption{\label{9}PCC for Nepali Audio Quality}
    \end{minipage}%
    \begin{minipage}{1.4\linewidth}
      \centering
      \small
        \begin{tabular}{lrrrrr}
                        & \multicolumn{1}{l}{R1} & \multicolumn{1}{l}{R2} & \multicolumn{1}{l}{R3} & \multicolumn{1}{l}{R4} & \multicolumn{1}{l}{R5} \\ \cline{2-6} 
\multicolumn{1}{l|}{R1} & 1                      & 0.19                   & 0.26                   & -0.02                  & -0.16                  \\
\multicolumn{1}{l|}{R2} & 0.19                   & 1                      & -0.05                  & 0.19                   & 0.22                   \\
\multicolumn{1}{l|}{R3} & 0.26                   & -0.05                  & 1                      & 0.39                   & 0.51                   \\
\multicolumn{1}{l|}{R4} & -0.02                  & 0.19                   & 0.39                   & 1                      & 0.59                   \\
\multicolumn{1}{l|}{R5} & -0.16                  & 0.22                   & 0.51                   & 0.59                   & 1                     
\end{tabular}
\caption{\label{10}PCC for Telugu Lip Synchronization}
    \end{minipage} 
\end{table}
\begin{table}[!htb]
    \begin{minipage}{0.9\linewidth}
      \centering
      \small
        \begin{tabular}{lrrrrr}
                        & \multicolumn{1}{l}{R1} & \multicolumn{1}{l}{R2} & \multicolumn{1}{l}{R3} & \multicolumn{1}{l}{R4} & \multicolumn{1}{l}{R5} \\ \cline{2-6} 
\multicolumn{1}{l|}{R1} & 1                      & -0.18                  & -0.18                  & 0.44                   & -0.13                  \\
\multicolumn{1}{l|}{R2} & -0.18                  & 1                      & 0.63                   & 0.31                   & 0.58                   \\
\multicolumn{1}{l|}{R3} & 0.39                   & 0.63                   & 1                      & 0.58                   & 0.67                   \\
\multicolumn{1}{l|}{R4} & 0.44                   & 0.31                   & 0.58                   & 1                      & -0.01                  \\
\multicolumn{1}{l|}{R5} & -0.13                  & 0.58                   & 0.67                   & -0.01                  & 1                     
\end{tabular}

\caption{\label{11}PCC for Telugu Translation Quality}
    \end{minipage}%
    \begin{minipage}{1.4\linewidth}
      \centering
      \small
        \begin{tabular}{lrrrrr}
                        & \multicolumn{1}{l}{R1} & \multicolumn{1}{l}{R2} & \multicolumn{1}{l}{R3} & \multicolumn{1}{l}{R4} & \multicolumn{1}{l}{R5} \\ \cline{2-6} 
\multicolumn{1}{l|}{R1} & 1                      & 0.83                   & 0.11                   & 0.32                   & 0.19                   \\
\multicolumn{1}{l|}{R2} & 0.83                   & 1                      & 0.59                   & 0.3                    & 0.63                   \\
\multicolumn{1}{l|}{R3} & 0.11                   & 0.59                   & 1                      & -0.19                  & 0.61                   \\
\multicolumn{1}{l|}{R4} & 0.32                   & 0.41                   & 0.3                    & 1                      & -0.08                  \\
\multicolumn{1}{l|}{R5} & 0.19                   & 0.63                   & 0.61                   & -0.08                  & 1                     
\end{tabular}
\caption{\label{12}PCC for Telugu Audio Quality}
    \end{minipage} 
\end{table}
\end{document}